\documentclass[a4paper,fleqn]{cas-dc}
\usepackage{ragged2e}
\usepackage{stfloats}
\usepackage{amsmath}
\usepackage{caption}
\usepackage[justification=centering]{caption}
\usepackage[numbers,sort&compress]{natbib}

\usepackage{graphicx}
\usepackage{epstopdf}
\usepackage{epsfig}
  
\def\tsc#1{\csdef{#1}{\textsc{\lowercase{#1}}\xspace}}
\tsc{WGM}
\tsc{QE}
\tsc{EP}
\tsc{PMS}
\tsc{BEC}
\tsc{DE}
\begin{document}
\let\WriteBookmarks\relax
\def\floatpagepagefraction{1}
\def\textpagefraction{.001}
\shorttitle{}
\shortauthors{Qingyuan Gong, Yu Liu, Liqiang Zhang,and Renhe Liu}

\title [mode = title]{Ghost-dil-NetVLAD: A Lightweight Neural Network for Visual Place Recognition}             

\tnotetext[1]{e-mail: gqy970506@tju.edu.cn; zhangliqiang@tju.edu.cn}
 \tnotetext[2]{This research is supported by National Natural Science Foundation of China (61771338).}

\author[1]{Qingyuan Gong}[style=chinese]
\address[1]{School of Microelectronics, Tianjin University, Tianjin, China}

\author[1]{Yu Liu}[style=chinese]

\author[1]{Liqiang Zhang}[style=chinese]
\cormark[1]
\cortext[cor1]{Corresponding author}

\author[1]{Renhe Liu}[style=chinese]

\begin{abstract}
Visual place recognition (VPR) is a challenging task with the unbalance between enormous computational cost and high recognition performance. Thanks to the practical feature extraction ability of the lightweight convolution neural networks (CNNs) and the train-ability of the vector of locally aggregated descriptors (VLAD) layer, we propose a lightweight weakly supervised end-to-end neural network consisting of a front-ended perception model called GhostCNN and a learnable VLAD layer as a back-end. GhostCNN is based on Ghost modules that are lightweight CNN-based architectures. They can generate redundant feature maps using linear operations instead of the traditional convolution process, making a good trade-off between computation resources and recognition accuracy. To enhance our proposed lightweight model further, we add dilated convolutions to the Ghost module to get features containing more spatial semantic information, improving accuracy. Finally, rich experiments conducted on a commonly used public benchmark and our private dataset validate that the proposed neural network reduces the FLOPs and parameters of VGG16-NetVLAD by $99.04\%$ and $80.16\%$, respectively. Besides, both models achieve similar accuracy.

\end{abstract}


\maketitle

\section{Introduction}

Visual place recognition (VPR) plays an incredible role in robotics and localization in recent years. VPR significantly contributes to the mission of identifying a single localization or simultaneous localization and mapping (SLAM) systems \cite{lowry2015visual}. Typically, The VPR problem can be formulated as image retrieval in an image database. To address the problem, a two-stage pipeline can be used: (1) extracting the vector representation from each image; (2) measuring the similarity between the query and each image in the database by a scoring function (e.g., Euclidean distance or cosine similarity) \cite{masone2021survey}.

Currently, the most effective feature extraction method (first step) in VPR is to use deep learning techniques. The original design goal of a convolutional neural network (CNN) is to create a network in which neurons in the first layer extracts local visual features, and neurons in the last layer fuse these features to form higher-order features. CNN can compress a large-size raw image to small-dimension data to retain practical image features. Generally, to obtain better performance, the number of layers of the network is constantly increasing \cite{hussain2018study}, from AlexNet \cite{krizhevsky2012imagenet} (7 layers) to VGG16 \cite{simonyan2014very} (16 layers), and then to ResNet \cite{he2016deep} with 152 layers.  

However, more convolution layers mean more computational resources and memories. Although the network performance can be improved, efficiency issues follow. Hence, various lightweight models emerge for simplifying the calculations and adapting the embedded devices. For efficiency issues, the usual method is to perform model compression. Compared with compressing a model, designing a lightweight model is a different approach: a more efficient network calculation method (mainly for the convolution method) is used to reduce model parameters with minor performance loss. For instance, GhostNet \cite{han2020ghostnet} utilized cheap operations (linear transformations) to generate feature maps instead of processing images by convolution operations, which could reveal important information underlying intrinsic features.   

Further, the critical point to improving VPR performance after feature extraction is to form the compact global image representation when subjected to various image transformations (second step) \cite{ku2015discriminatively,arandjelovic2016netvlad}. VPR mission aggregates the extracted local feature descriptors to the global feature representations. The global descriptors \cite{zhang2021visual} can be aggregated by bag-of-word (BoW), Fisher Vector (FV) and vector of locally aggregated descriptors (VLAD), etc. BoW uses the nearest neighbour cluster centers to represent feature points. FV clusters the data by Gaussian mixture model and then use the linear combination of all cluster centers to approximate feature points. VLAD is a popular descriptor pooling method, for instance, level retrieval \cite{zhou2017recent,kapoor2021state}. It captures information about the statistics of local descriptors aggregated over the image and stores the sum of residuals (difference vector between the descriptor and its corresponding cluster center) for each visual word. However, features extracted by these methods are all handcrafted and not robust enough to the environmental changes such as lighting conditions, scales and viewpoints. Inspired by VLAD, a novel CNN-based architecture, VGG16-NetVLAD, was design to mimic it \cite{arandjelovic2016netvlad}. It regards the output (i.e., $K \times W \times H$ feature map) of the front-ended CNN as $W \times H$ local descriptors with the length of $K$. It aggregates them with a specifically designed pooling layer to obtain the final global representation. Experimental results show its excellent performance on challenging public VPR dataset such as Pitts30k \cite{torii2013visual}. NetVLAD architecture is also applied in other image retrieval area. GhostVLAD is used in template-based face recognition task \cite{2018GhostVLAD}. It adds ghost clusters to NetVLAD for higher accuracy. GM-NetVLAD \cite{9167393} model is used in social media platforms for image retrieval task. The model constructs feature pyramid network based on ResNet backbone and computes NetVLAD features at each pyramid level to capture the multi-scale information to improve social media image retrieval performance. However, the computational cost of the traditional CNN-NetVLAD architecture is considerable.

Our main contributions in designing a VPR model with fast computational speed and high accuracy can be summarized as follows:
\begin{itemize}
\item we design a lightweight VPR model named Ghost-dil-NetVLAD with dilated convolution-based Ghost modules and a learn-able VLAD layer;

\item we establish and open a location dataset of the Tianjin University campus named TJU-Location Dataset, including images of 50 iconic locations. In addition, two public datasets named Pitts30k and Tokyo 24/7 are also used to discuss the performance of the proposed model and several VPR models.
\end{itemize}

The remaining framework of this paper is as follows: related works are presented in Section II. In Section III, we propose a lightweight model named Ghost-dil-NetVLAD. Section IV contains the experimental results and the ablation experiments of different fusion methods. The last section is the conclusion.

\section{Related work}
\subsection{Lightweight VPR Model}
Feature extraction is the first step in the VPR mission. The traditional feature representations include scale-invariant feature transform (SIFT), FAB-MAP and Cross-Region-BoW, improving the robustness against appearance or illumination changes. However, these methods always lead to a large amount of calculation compared to feature extracted by deep learning methods. Khaliq \emph{et al.} \cite{khaliq2019holistic,khaliq2019camal} propose two light\-weight VPR methods: CAMAL and Region-VLAD. The CAMAL captures multi-layer context-aware attentions robust under changing environments and viewpoints. The Region-VLAD employs middle convolutional layers of AlexNet365 to process regional vocabulary and extraction of VLAD for image matching. Although these methods realize lightweight structures, the convolutional structure of the front-end CNN is not optimized. Therefore, it still leads to considerable computational cost.
\subsection{Deep Learned Representations for VPR}
For VPR tasks, CNNs are specialized for processing images or videos \cite{zhou2017recent}. With the excellent performance of CNNs in the field of vision classification and segmentation in recent years, it has been shown that the image representations generated by CNNs can be transferable to other visual tasks. In other words, the feature extracted from CNNs can also be applied to image retrieval tasks to achieve good results. Namely, the feature extracted by CNNs is robust and discriminative in changing environments \cite{krizhevsky2012imagenet}. Babenko \emph{et al.} \cite{babenko2014neural} propose that the feature map with a size of $H \times W \times C$ generated by CNNs for image classification obtains impressive results as representations in the VPR task. Simply flattening the feature maps of a convolution layer does not take full advantage of the spatial information \cite{goodfellow2016deep}. Therefore, improving the current state-of-the-art representations for visual place recognition can add aggregated representations layers behind the feature maps.

The aggregated representation layers can be seen as an $H \times W$ grid of $C$-dimensional feature descriptors to a vector representing the original image, aggregating local features into a representative global feature. And then, a similarity function (e.g., Euclidean distance or cosine similarity) is leveraged to select the best candidate for image retrieval. There are many classic algorithms about image vector aggregating, e.g., VLAD \cite{paulin2015local}, BoW \cite{mohedano2016bags}, Fisher vector \cite{sanchez2013image} and aggregated selective matching kernel (ASMK) \cite{cao2020unifying}. However, these algorithms are not robust enough to environmental changes such as lighting conditions, scales and viewpoints. Thanks to the development of deep learning technology, a model named NetVLAD implements the VLAD layer embedding and aggregation behind the CNN structure for VPR \cite{arandjelovic2016netvlad}. In recent years, \cite{yu2019spatial,ge2020self,hausler2021patch,xu2021esa} improve the origin NetVLAD model with the process of changing the loss function or adding feature matching to enhance the performance. Among these models, Patch-NetVLAD \cite{hausler2021patch} achieves the state-of-the-art performance on Pitts30k \cite{torii2013visual} and Tokyo 24/7 \cite{201524} datasets. Nevertheless, none of them considers the computing power required by the actual operation of the model and the potentiality of these models adapted to embedded devices.

\subsection{Lightweight CNN}
There are two main methods of lightening neural networks: model compression and compact model design. Mo\-del compression includes binarization methods, tensor decomposition, and knowledge distillation, etc. They can accelerate the model by efficient binary operations, exploit the redundancy and low-rank property in parameters, or utilize larger models to teach smaller ones. In addition, these methods can not fundamentally improve the lightweight degree of the model. The improvement on basic operations and architectures will make them go further \cite{han2020ghostnet}. 

There are many lightweight CNN models by a compact model design, like MobileNets and ShuffleNet, etc. MobileNetV1 \cite{howard2017mobilenets} is first proposed by Google. It is of small size and low calculation, suitable for mobile devices. It is lightweight, using depthwise separable convolution instead of standard convolution, and uses width multiply to reduce parameters. MobileNetV2 \cite{sandler2018mobilenetv2} proposes inverted residual block, and MobileNetV3 \cite{howard2019searching} further utilizes AutoML technology, achieving better performance with fewer floating-point operations (FLOPs). ShuffleNet \cite{zhang2018shufflenet} is a lightweight network structure proposed by Face++. The main thought is to use Group convolution and Channel shuffle to improve ResNet, regarded as a compressed version of ResNet. ShuffleNetV2 \cite{ma2018shufflenet} introduces channel split operation to realize fast and high precision. GhostNet \cite{han2020ghostnet} uses a series of simple linear transformations with cheap cost to generate ghost feature maps that could fully reveal information underlying intrinsic features. In efficiency and accuracy, GhostNet is better than others. In the work of GhostNet, the Ghost module is introduced and can be taken as a plug-and-play component.
\begin{figure*}
  \centering
     \includegraphics[scale=0.27]{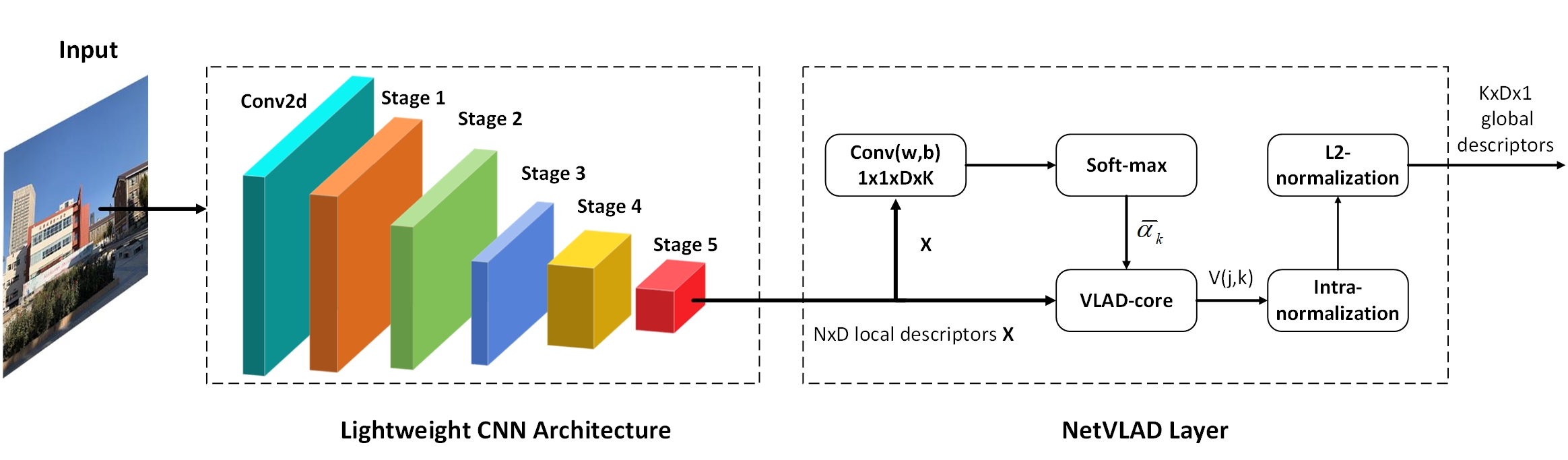}
  \caption{The proposed framework for place recognition. The feature extraction part is lightweight CNN architecture, which is a 5-stage CNN architecture. Through every stage, the feature map becomes a half one. The NetVLAD layer processes local descriptors to global descriptors.}
  \label{figure1}
\end{figure*}

\section{Methodology}\label{sec:approach}
The framework of our proposed Ghost-dil-NetVLAD is shown in Fig. \ref{figure1}. The Ghost-dil-NetVLAD contains two parts. One is the lightweight feature extraction architecture (GhostCNN) shown in Section 3.1, and the remaining is the NetVLAD layer described in Section 3.2.  
\begin{table}[t]
\centering
\caption{GhostCNN architecture.}
\label{tab1}
\setlength\tabcolsep{2.5 pt}
\begin{tabular}{c|c|c|c|c}
\hline
\multirow{2}{*}{Stage}   & \multirow{2}{*}{output size} & \multirow{2}{*}{Operator}         & \multirow{2}{*}{Stride} & \multirow{2}{*}{SE} \\&&                                   &                         &                     \\ \hline
-                        & 320x240x24                   & Conv2d 3x3                        & 2                       & 0                   \\ \hline
\multirow{4}{*}{stage1}  & \multirow{4}{*}{160x120x40}  & \multirow{2}{*}{Ghost bottleneck} & \multirow{2}{*}{1}      & \multirow{2}{*}{0}  \\
                         &                              &                                   &                         &                     \\ \cline{3-5} 
                         &                              & \multirow{2}{*}{Ghost bottleneck} & \multirow{2}{*}{2}      & \multirow{2}{*}{0}  \\
                         &                              &                                   &                         &                     \\ \hline
\multirow{4}{*}{stage2}  & \multirow{4}{*}{80x60x112}   & \multirow{2}{*}{Ghost bottleneck} & \multirow{2}{*}{1}      & \multirow{2}{*}{0}  \\
                         &                              &                                   &                         &                     \\ \cline{3-5} 
                         &                              & \multirow{2}{*}{Ghost bottleneck} & \multirow{2}{*}{2}      & \multirow{2}{*}{1}  \\
                         &                              &                                   &                         &                     \\ \hline
\multirow{4}{*}{stage3}  & \multirow{4}{*}{40x30x160}   & \multirow{2}{*}{Ghost bottleneck} & \multirow{2}{*}{1}      & \multirow{2}{*}{1}  \\
                         &                              &                                   &                         &                     \\ \cline{3-5} 
                         &                              & \multirow{2}{*}{Ghost bottleneck} & \multirow{2}{*}{2}      & \multirow{2}{*}{0}  \\
                         &                              &                                   &                         &                     \\ \hline
\multirow{12}{*}{stage4} & \multirow{12}{*}{20x15x960}  & \multirow{2}{*}{Ghost bottleneck} & \multirow{2}{*}{1}      & \multirow{2}{*}{0}  \\
                         &                              &                                   &                         &                     \\ \cline{3-5} 
                         &                              & \multirow{2}{*}{Ghost bottleneck} & \multirow{2}{*}{1}      & \multirow{2}{*}{0}  \\
                         &                              &                                   &                         &                     \\ \cline{3-5} 
                         &                              & \multirow{2}{*}{Ghost bottleneck} & \multirow{2}{*}{1}      & \multirow{2}{*}{0}  \\
                         &                              &                                   &                         &                     \\ \cline{3-5} 
                         &                              & \multirow{2}{*}{Ghost bottleneck} & \multirow{2}{*}{1}      & \multirow{2}{*}{1}  \\
                         &                              &                                   &                         &                     \\ \cline{3-5} 
                         &                              & \multirow{2}{*}{Ghost bottleneck} & \multirow{2}{*}{1}      & \multirow{2}{*}{1}  \\
                         &                              &                                   &                         &                     \\ \cline{3-5} 
                         &                              & \multirow{2}{*}{Ghost bottleneck} & \multirow{2}{*}{2}      & \multirow{2}{*}{1}  \\
                         &                              &                                   &                         &                     \\ \hline
\multirow{8}{*}{stage5}  & \multirow{8}{*}{20x15x960}   & \multirow{2}{*}{Ghost bottleneck} & \multirow{2}{*}{1}      & \multirow{2}{*}{0}  \\
                         &                              &                                   &                         &                     \\ \cline{3-5} 
                         &                              & \multirow{2}{*}{Ghost bottleneck} & \multirow{2}{*}{1}      & \multirow{2}{*}{1}  \\
                         &                              &                                   &                         &                     \\ \cline{3-5} 
                         &                              & \multirow{2}{*}{Ghost bottleneck} & \multirow{2}{*}{1}      & \multirow{2}{*}{0}  \\
                         &                              &                                   &                         &                     \\ \cline{3-5} 
                         &                              & \multirow{2}{*}{Ghost bottleneck} & \multirow{2}{*}{1}      & \multirow{2}{*}{1}  \\
                         &                              &                                   &                         &                     \\ \hline
\end{tabular}
\end{table}
\subsection{GhostCNN}
For the traditional CNN, the redundancy in feature maps always guarantees a comprehensive understanding of the input data. For example, there are many similar feature maps through the convolution layer of ResNet \cite{he2016deep}, just like ghosts of each other. The redundancy in feature maps is an essential feature for the success of deep neural networks. However, it results in massive computational costs. Inspired by basic Ghost modules in the GhostNet \cite{han2020ghostnet}, we design the lightweight neural network named GhostCNN for front-ended feature extraction. Ghost modules skillfully utilize linear transformations to generate ghost feature map pair examples to reduce the computational cost significantly and ensure a satisfactory accuracy simultaneously.
The proposed GhostCNN consists of a convolution operation and five stages. After each step, the size of the output feature maps are reduced by half till the stage 4. There are at least two Ghost bottlenecks \cite{han2020ghostnet} in each stage, and each bottleneck contains two Ghost modules (its architecture is shown in Fig. \ref{figure3}). The detailed architecture of the GhoseCNN is given in Table 1.

\begin{figure*}[t]
\centering
\includegraphics[scale=0.3]{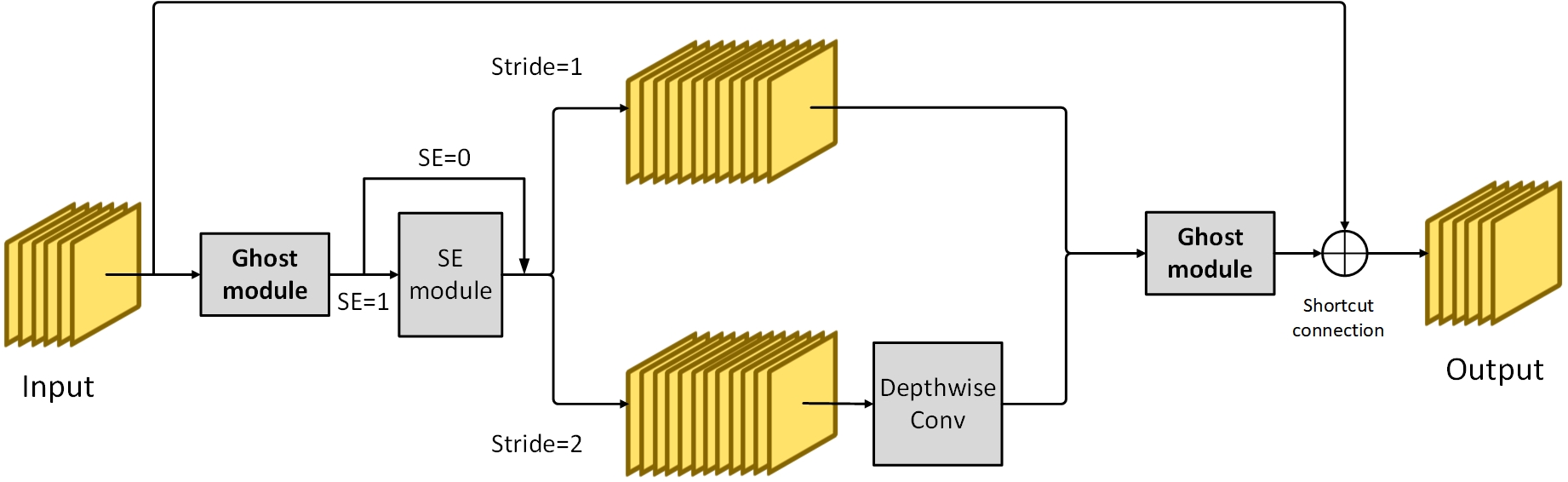}
\caption{Ghost bottleneck architecture.}
\label{figure2}
\end{figure*}
\subsubsection{Ghost Bottleneck}
Similar to the basic residual block in ResNet, the Ghost bottleneck consists of two stacked Ghost modules, as shown in Fig. \ref{figure2}. The first Ghost module is used as an expansion layer, increasing the number of channels, and the second Ghost module reduces the number of channels to match the shortcut. Then, the input and the result from the second Ghost module are fed into the shortcut connection to generate the final output. If Stride=2, a depthwise convolution layer is inserted between the two Ghost modules. Depthwise convolution processes the image from each channel simultaneously, and the number of feature maps after this operation is the same as the number of input channels. If SE=1, the squeeze-and-excitation (SE) module is selected. The SE block \cite{2018Squeeze} comprises an
average pooling layer and two pointwise convolutions. It is implemented to enhance the channelwise feature responses. 

\subsubsection{Ghost Module}
First, given the input data $X \in \mathbb{R}^{c \times h \times w}$, $m$ feature maps ($Y\in \mathbb{R}^{h \times w \times m}$ where $h$ and $w$ are the height and width of the input data, respectively) can be generated using
\begin{equation}
\label{eq1}
\mathop{Y=X * f}
\end{equation} 
where $f \in \mathbb{R}^{c \times k \times k \times m}$ is the convolution filter with the kernel size of $k \times k$; $c$ is  the number of input channels. For the traditional CNN, the original number of intrinsic feature maps is $D$. But after this step, the number of output feature maps is $m$ ($m \leq D$).
 
Second, cheap operation (i.e., Eq. \ref{eq2}) including identity transformation is performed on each feature map in $Y$ to obtain the desired $n$ feature maps. Each feature map in $Y$ generates $s$ ghost feature maps, a total of $m \times s$, to ensure that the feature map shape of the Ghost module and the standard convolution output is the same. 
\begin{equation}
\label{eq2}
\mathop{y_{i j}=\Phi_{i, j}\left(y_{i}\right), \quad \forall i=1, \ldots, m, \quad j=1, \ldots, s}
\end{equation}
where $y_{i}$ is the $i$-th intrinsic feature map in $Y$, $\Phi_{i, j}$ is the $j$-th linear operation (except the last one) to generate the $j$-th ghost feature map $y_{i j}$. Namely, $y_{i}$ can have one or more ghost feature maps $\left\{y_{i j}\right\}_{j=1}^{s}$. The last $\Phi_{i, s}$ is the identity mapping for preserving the intrinsic feature maps. Through Eq. \ref{eq2}, we have $D=m \times s$ feature maps. $Y'=\left[y_{11}, y_{12}, \cdots, y_{m s}\right]$ as the output data of a Ghost module.
\begin{figure}
\centering
\includegraphics[scale=0.26]{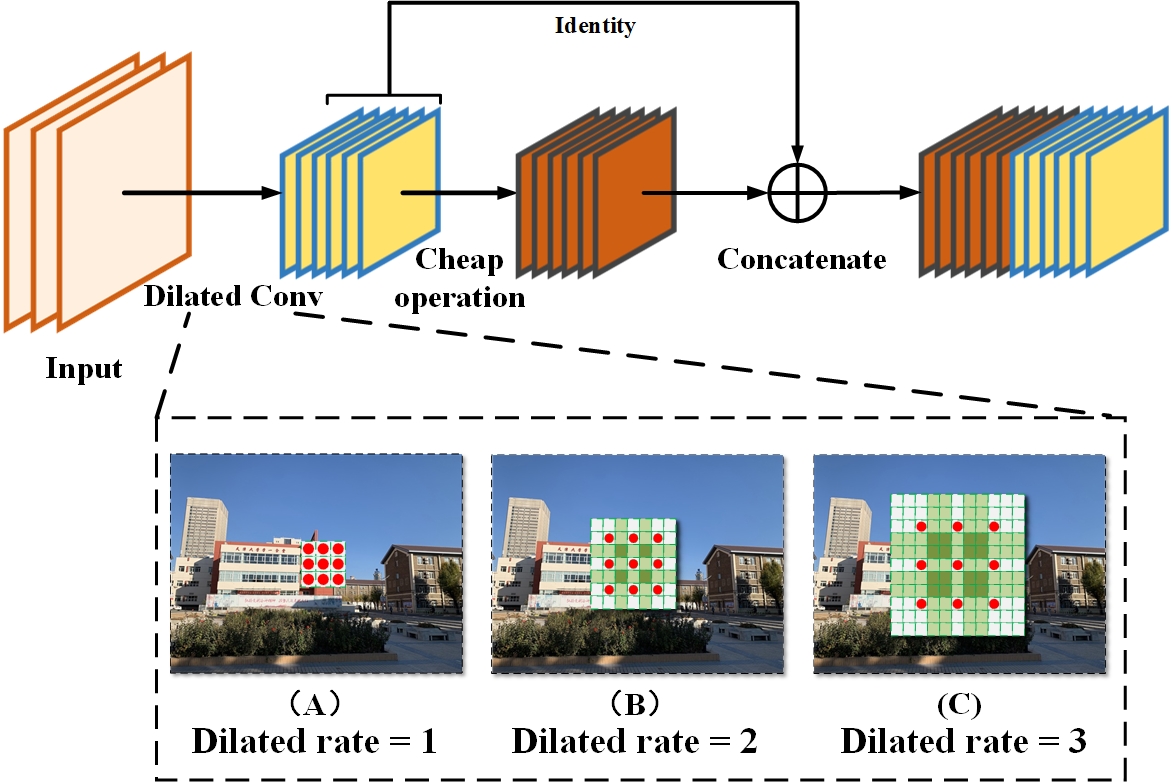}
\caption{Ghost module with dilated convolution filter. We add dilated convolutions to the first step of the Ghost module. Sub-figure (A) shows the $3 \times 3$ receptive field of 1-dilated convolution filter (i.e., ordinary convolution). (B) reveals the $7 \times 7$ receptive field of 2-dilated convolution filter. (C) indicates the $11 \times 11$ receptive field of 3-dilated convolution filter.}
\label{figure3}
\end{figure}

From Fig. \ref{figure3}, the dilated convolution part shows the different receptive fields of convolution filter with different dilated rates. For different dilated rates, the number of parameters associated with each layer is identical. Intuitively, when multiple convolution kernels with different dilation rates are superimposed, other receptive fields will bring multi-scale context information. Therefore, we add dilated convolutions in the Ghost module to improve the model performance. Dilated convolutions support an exponential expansion of the receptive field, which means that the output feature maps contain more global features and semantically higher-level features. It is conducive to the image recognition task.

\subsection{NetVLAD layer} 
Vector of locally aggregated descriptors (VLAD) \cite{jegou2010aggregating} store the sum of residuals (difference vector between the descriptor and its corresponding cluster center) for each visual word. The NetVLAD \cite{arandjelovic2016netvlad} uses the CNN architecture to capture the information about the statistics of local descriptors aggregated over the image. Given the input image $I$, the local descriptor $Y'$ can be obtained by
\begin{equation}
\label{local_descriptors}
f_{GhostCNN}: I \rightarrow Y' \in \mathbb{R}^{h \times w \times D}
\end{equation}
In other words, the output of GhostCNN's last convolution layer is a $h \times w \times D$ feature map which can be considered as a set of $D$-dimensional descriptors extracted at $h \times w$ spatial locations. Similarly, the feature map can be deemed as $N$ ($N=h \times w$) $D$-dimensional feature descriptors $\{\mathbf{x}_i\}_{i=1}^N$ with each of them representing the local features at specific local positions of the input image.

Formally, given local image descriptors $\{\mathbf{x}_i\}_{i=1}^N$ as input, and cluster centers (visual words) $\{\mathbf{c}_i\}_{i=1}^K$ as VLAD parameters, the output VLAD image representation $V$ is $D \times K$-dimensional. The $\left(j,k\right)$ element of $V$ can be expressed by
\begin{equation}
\label{eq4}
\mathop{V(j, k)=\sum_{i=1}^{N} \bar{\alpha}_{k}\left(\mathbf{x}_{i}\right)\left(x_{i}(j)-c_{k}(j)\right)}
\end{equation}
where $x_{i}(j)$ is the $j$-th dimensions of the $i$-th descriptor and $c_k(j)$ denotes the $k$-th cluster center; $\bar{\alpha}_{k}\left(\mathbf{x}_{i}\right)$ is the weight between the descriptor $\mathbf{x}_{i}$ and the $k$-th cluster center. The weight ranges from 0 to 1, with the highest weight assigned to the closest cluster center. Namely, $\bar{\alpha}_{k}\left(\mathbf{x}_{i}\right)$ is 1 if cluster $\mathbf{c}_k$ is closest to the descriptor $\mathbf{x}_{i}$ and 0 otherwise. $\sum_{i=1}^{N} \bar{\alpha}_{k}\left(\mathbf{x}_{i}\right)=1$. The weight is trainable via back-propagation and can be described as follows
\begin{equation}
\label{eq5}
\mathop{\bar{\alpha}_{k}\left(\mathbf{x}_{i}\right)=\frac{e^{\alpha\left\|\mathbf{x}_{i}-\mathbf{c}_{k}\right\|^2}}{\sum_{k^{\prime}} e^{-\alpha\left\|\mathbf{x}_{i}-\mathbf{c}_{k'}\right\|^2 }}}
\end{equation}
which assigns the weight of descriptor $\mathbf{x}_i$ to cluster $\mathbf{c}_k$ according to their proximity.

For convenience, the output VLAD image representation $V$ is written as a $D \times K$ matrix, and this matrix is reshaped into a row vector. After normalization, we have a $\left( K \times D\right) \times 1$ image representation. Then, the dimension reduction is performed using principal component analysis (PCA) with whitening followed by L2-normalization \cite{jegou2012negative}.

\section{Experiments}\label{sec:Results}
This section describes our used datasets (Section 4.1) and evaluation metrics (Section 4.2). Finally, we perform quantitative and qualitative experiments to validate our proposed approach (Section 4.4).

\subsection{Datasets}

First, we leverage a commonly used public dataset (\emph{Pitts30k-train}) to train our model. Then, two publicly available datasets (\emph{Pitts30k-test} and \emph{Tokyo 24/7}) and our established dataset (\emph{TJU-Location}) are used to verify our method.

\begin{figure}[t]
\centering
\includegraphics[scale=0.23]{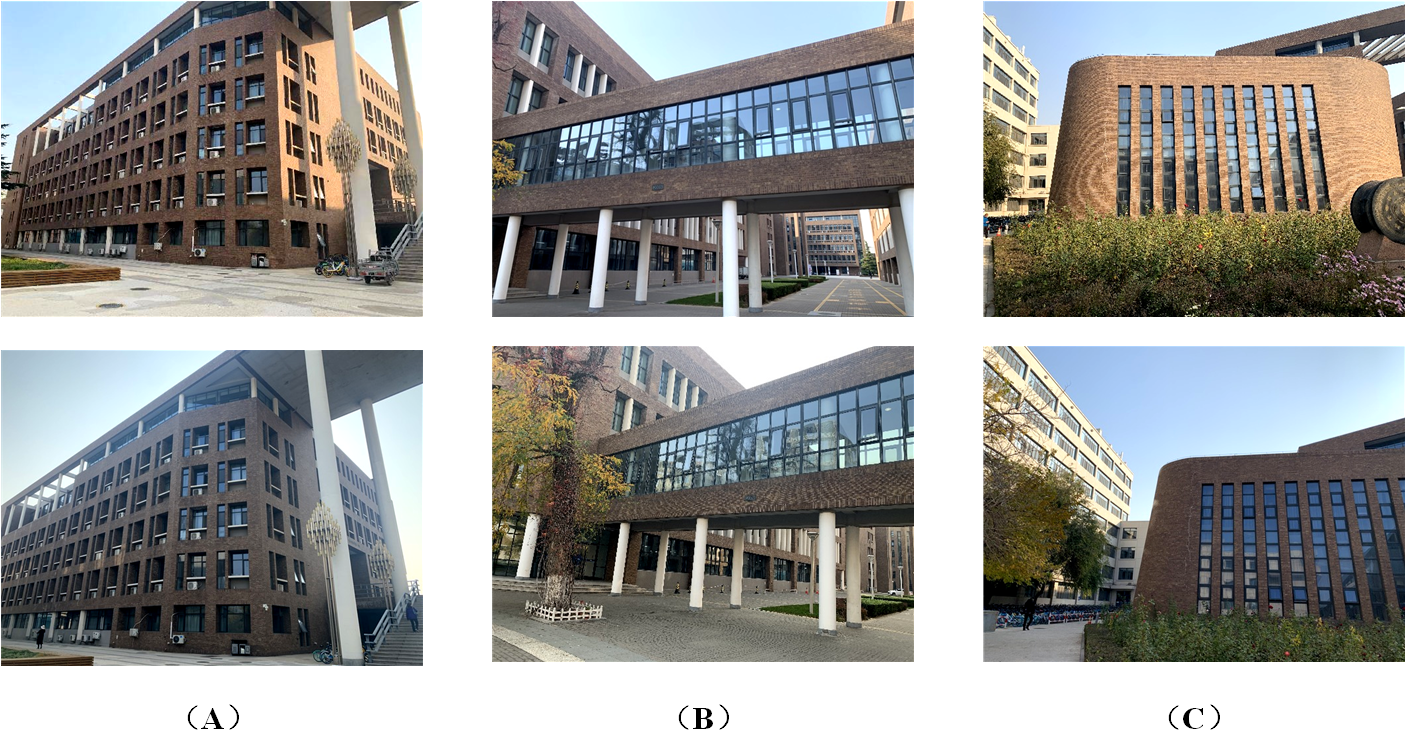}
\caption{TJU-Location dataset examples. Each column shows perspective images taken at different times.}
\label{figure4}
\end{figure}

\begin{itemize}
\item \emph{Pitts30k} \cite{arandjelovic2016netvlad,torii2013visual}: It is a public dataset containing 30 k database images downloaded from Google Street View and 22 k test queries captured at different times. The dataset is divided into three roughly equal parts for training, validation, and testing. Street views in the three sets are geographically disjoint. Each set contains 10 k images to process and 8 k queries, and the queries are geographically disjoint with the processed images.

\item \emph{Tokyo 24/7} \cite{201524}: It is a public test dataset containing 76 k database images and 315 query images taken by mobile phone cameras. This is an extremely challenging dataset where the queries were taken at daytime, sunset and night, while the database images were only taken at daytime from Google Street View.

\item \emph{TJU-Location}: As shown in Fig. \ref{figure4}, our dataset contains multiple street-level images taken at different times. We selected 50 positions with distinctive characteristics of the buildings on the Weijin campus of Tianjin University.  All points are isolated geographically. We recorded images from these locations by two smartphones (i.e., iPhone XR and Huawei Mate 20). In one place, images were token per 45 degrees in the horizontal plane. In one horizontal direction, two images with different pitch angles were captured. We collected images in one location at 9:00 and 14:00, respectively. Therefore, 64 images were recorded at each location. Finally, a database containing 3.2 k images with 2.3 k queries was established. We split all database images into three roughly equal parts for training, validation, and testing. These subsets were independent of each other geographically. To obtain accurate positions of the collected images, we used the portable positioning module called Qianxun magic cube MC120M with sub-meter accuracy from Qianxun spatial intelligence company. We uploaded entire data\-set to Baidu cloud disk and opened it. \par (\emph{https://pan.baidu.com/s/1F-uLbLUiD2rCmsXcN73uFw} and the code is \emph{8wyd})

\end{itemize}
\subsection{Evaluation Metrics} 

For all the datasets, they are evaluated using the Recall@N. The Recall quantifies the number of positive class predictions made out of all positive examples in the dataset. That is, the correct prediction is the proportion of all positive samples. The query image is correctly localized if at least one of the top N images is within the ground-truth tolerance. The queries are correctly recognized (positive proportion) if database images are retrieved within a specific range from true positions (GNSS positions).

In addition, model size and computing power requirements are essential indexes to evaluate the potential of deep learning models to be deployed in embedded systems. Therefore, two metrics, model parameters and FLOPs are introduced. Model parameters are used to reveal the model's complexity, and FLOPs reflect requirements for hardware such as GPU. 

\subsection{Implementation Details} 

All models were implemented on the publicly available PyTorch framework. Two NVIDIA Tesla V100 GPUs were leveraged for training, validation, and testing. In addition, we used stochastic gradient descent (SGD) as an optimizer. Its starting learning rate is $1 \times 10 ^ {- 4} $, and the parameter ``weight \_decay" is $1 \times 10 ^ {- 3} $. The batch size is 4 and the momentum is 0.9. For preprocessing, all images were resized to $640\times480$ pixels.

We used three representative CNN architectures (i.e., Alex\-Net, VGG-16 and MobileNetV3) and our proposed GhostCNN as front-ends to extract feature maps. AlexNet and VGG-16 are cropped at the last convolution layer (conv5) with the encoder dimension (i.e., $D$-dimension) of 256 and 512 before ReLU, respectively. MobileNetV3 is cropped at the last stage with the encoder dimension of 960 before adaptive average pooling layer. 

To train high-performance models, we used the triplet loss \cite{arandjelovic2016netvlad} as the loss function. The function consists of a triplet tuple $\left(q,\left\{p_{i}^{q}\right\},\left\{n_{j}^{q}\right\}\right)$, where $\left\{p_{i}^{q}\right\}$ is a set of potential positives (geographical distance within a specific range) and $\left\{n_{j}^{q}\right\}$ is a set of definite negatives (geographical distance a specific range). First, we need to choose the best positive ($p_{i *}^{q}={\operatorname{argmin}} d_{\theta}\left(q, p_{i}^{q}\right)$). Namely, for a given test query image $q$, we wish that the euclidean distance between the query $q$ and the best potential positive to be smaller than its distance to definite negative, which is described as
\begin{equation}
\label{eq6}
\mathop{d_{\theta}\left(q, p_{i *}^{q}\right)<d_{\theta}\left(q, n_{j}^{q}\right), \forall j}
\end{equation}
Through the above inequality relationship, the loss function $L_{\theta}$ can be described as:
\begin{equation}
\label{eq7}
\mathop{L_{\theta}=\sum_{j} l\left(\min _{i}\left(  d_{\theta}^{2}\left(q, p_{i}^{q}\right)\right) +m-d_{\theta}^{2}\left(q, n_{j}^{q}\right)\right)}
\end{equation}
where $m$ is a constant parameter as 0.1 to ensure that the query is close to the potential positive and away from the definite negative; $l$ is the hinge loss $l\left(x \right)=max\left(0,x \right)  $.

\emph{Pre-train} In NetVLAD \cite{arandjelovic2016netvlad}, VGG-16 and AlexNet are pre-trained by ImageNet \cite{imagenet_cvpr09} for classification, which demonstrates that the end-to-end model achieves higher accuracy and training speed \cite{he2019rethinking,hendrycks2019using}. The ImageNet dataset has more than 14 million images, covering more than 20000 categories, and it is widely used for deep learning perception. Nevertheless, most categories are objects and creatures. However, these images lack the point, line and angle features in the building images. Since we live in various buildings, more building features help improve the VPR model performance. Therefore, we pre-train all models on the Places-365 dataset \cite{zhou2017places} to ensure that they are sensitive to building features. Places-365 is a dataset for building classification and scene recognition. It has more than 18 million training images, with the image number per class varying from 3068 to 5000 for 365 categories. The validation set has 50 images per class, and the test set has 900 images per class.  
\subsection{Results and Discussion}
In this section, six models including Alex-NetVLAD, VGG16-NetVLAD, Patch-NetVLAD (Considering our limited computational resources, we only use its built-in storage mode. In this paper, we uniformly call this method Patch-NetVLAD.), MobileNetV3-NetVLAD (lightweight CNN + NetVLAD), Ghost-NetVLAD (the Ghost module does not have dilated convolutions) and Ghost-dil-NetVLAD (the Ghost module has dilated convolutions) are trained on the Pitts30k-train dataset, and evaluated on the Pitts30k test, Tokyo 24/7 and TJU-Location test datasets. Sufficient experiments are performed to discuss different model performances.

\subsubsection{Model Accuracy}

\begin{table*}
	\centering
	\caption{Recall@N of Alex-NetVLAD, VGG16-NetVLAD, Patch-NetVLAD, MobileNetV3-NetVLAD, Ghost-NetVLAD and Ghost-dil-NetVLAD on the Pitts30k test dataset, Tokyo 24/7 and TJU-Location test dataset. We report all results for each of them, including the {\color{red}best}, {\color{blue}second-best} and {\color{orange}third-best} results.}
	\label{tab2}
	\begin{tabular}{ccccccc}
\hline
\textbf{Dataset}                         & \textbf{Model}               & \textbf{recall@1} & \textbf{recall@5} & \textbf{recall@10} & \textbf{recall@20} & \textbf{recall@25} \\ \hline
                                & Alex-NetVLAD        & 69.87                           & 85.02                           & 89.14                            & 92.52                            & 93.81                            \\
                                & VGG16-NetVLAD       & {\color{blue}80.65}                           & {\color{blue}90.55}                           & {\color{blue}93.38}                            & {\color{blue}95.47}                            & {\color{blue}95.97}                            \\
                                & Patch-NetVLAD       & {\color{red}87.02}                           & {\color{red}92.02}                           & {\color{red}94.78}                            & {\color{red}95.79}                            & {\color{red}96.08}                            \\
                                & MobileNetv3-NetVLAD & 73.90                           & 86.59                           & 90.67                            & 93.87                            & 94.50                            \\
                                & Ghost-NetVLAD    & 76.38                           & 87.88                           & 91.34                            & 93.90                            & 94.72                            \\
\multirow{-6}{*}{Pitts30k test} & Ghost-dil-NetVLAD   & {\color{orange}79.45}                           & {\color{orange}89.67}                           & {\color{orange}92.80}                            & {\color{orange}95.35}                            & {\color{orange}95.95}                            \\ \hline
                                & Alex-NetVLAD        & 69.66                           & {\color{blue}79.04}                           & {\color{orange}84.11}                            & {\color{blue}88.15}                            & {\color{orange}89.19}                            \\
                                & VGG16-NetVLAD       & {\color{red}72.01}                           & {\color{blue}79.04}                           & 82.42                            & {\color{blue}88.15}                            & {\color{blue}89.32}                            \\
                                & Patch-NetVLAD       & {\color{blue}71.22}                           & {\color{orange}78.26}                           & {\color{blue}84.64}                            & {\color{orange}88.02}                            & {\color{blue}89.32}                            \\
                                & MobileNetv3-NetVLAD & 69.01                           & 77.08                           & 81.77                            & 86.59                            & 88.15                            \\
                                & Ghost-NetVLAD    & 67.97                           & 76.43                           & 81.51                            & 86.33                            & 87.24                            \\
\multirow{-6}{*}{TJU-Location test}       & Ghost-dil-NetVLAD   & {\color{orange}70.83}                           & {\color{red}80.08}                           & {\color{red}85.29}                            & {\color{red}89.19}                            & {\color{red}89.84}                            \\ \hline
                                & Alex-NetVLAD        & 40.32                           & 51.43                           & {\color{orange}61.27}                            & {\color{orange}68.25}                            & {\color{orange}68.89}                            \\
                                & VGG16-NetVLAD       & {\color{blue}56.19}                           & {\color{blue}64.13}                           & {\color{blue}74.29}                            & {\color{red}80.32}                            & {\color{red}81.59}                            \\
                                & Patch-NetVLAD       & {\color{red}66.35}                           & {\color{red}69.52}                           & {\color{red}75.24}                            & {\color{blue}76.51}                            & {\color{blue}77.14}                            \\
                                & MobileNetv3-NetVLAD & 37.46                           & 46.67                           & 53.33                            & 58.10                            & 60.00                            \\
                                & Ghost-NetVLAD    & 34.92                           & 40.63                           & 52.06                            & 57.78                            & 59.37                            \\
\multirow{-6}{*}{Tokyo 24/7}    & Ghost-dil-NetVLAD   & {\color{orange}41.90}                           & {\color{orange}54.29}                           & 57.46                            & 60.32                            & 62.86                            \\ \hline
\end{tabular}
\end{table*}


Table 2 shows the Recall@N of six models on the Pitts30k test dataset, TJU-Location test dataset and 
Tokyo 24/7 dataset. The experimental results demonstrate that Patch-NetVLAD achieves the best performance on the Pitts30k test dataset and Tokyo 24/7 dataset, while Ghost-dil-NetVLAD performs the best on TJU-Location test dataset because most Recall@N of Ghost-dil-NetVLAD are greater than those of the remaining models. VGG16-NetVLAD and Ghost-dil-NetVLAD achieve closed performance on the Pitts30k test dataset and Tokyo 24/7 dataset. By contrast, Patch-NetVLAD and VGG16-NetVLAD achieve closed accuracy on the Tokyo 24/7 dataset. Their performance is better than that of all lightweight models (i.e., MobileNetv3-NetVLAD, Ghost-NetVLAD and Ghost-dil-NetVLAD). The Tokyo 24/7 dataset is challenging because the queries are taken at sunset or night and need to be retrieved from the daytime database, which means the tested model should be robust enough to illumination changing. For the standard CNN architecture, more standard convolutional calculations could extract more redundant features. This mechanism can improve the accuracy of image recognition under low-light condition. It may be the reason why all lightweight models perform worse than VGG16-NetVLAD and Patch-NetVLAD on the Tokyo 24/7 dataset. 
Among the three lightweight models, Ghost-dil-NetVLAD performs the best. The possible reason is that Ghost-dil-NetVLAD leverages cheap operations to simplify standard convolutional calculations. Essentially, compared with the lightweight MobileNetV3, GhostCNN preserves more redundancy in feature maps. Meanwhile, expanded receptive field makes feature maps with more interrelated characteristics.

\subsubsection{Model Efficiency}

\begin{figure*}
	\centering
	\includegraphics[scale=0.5]{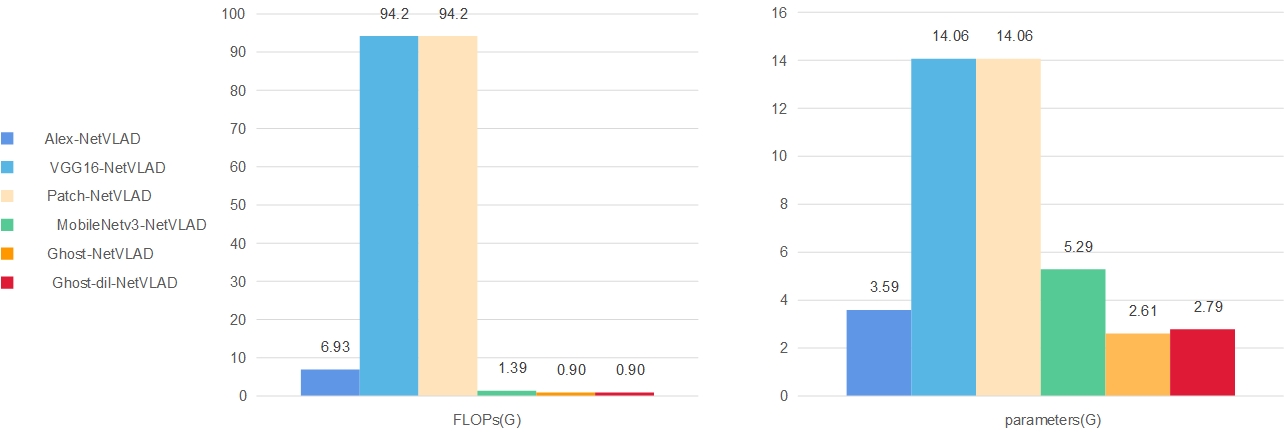}
	\caption{FLOPs and parameters of the Alex-NetVLAD, VGG16-NetVLAD, Patch-NetVLAD, MobileNetV3-NetVLAD, Ghost-NetVLAD and Ghost-dil-NetVLAD.}
	\label{figure5}
\end{figure*}

As shown in Fig. \ref{figure5}, it is evident that Ghost-NetVLAD outperforms other models in FLOPs and model parameters. There are few differences between  Ghost-dil-NetVLAD and Ghost-NetVLAD in the model efficiency. However, VGG16-NetVLAD and Patch-NetVLAD need the most computational resources, which is not conducive to its deployment in embedded devices. Compared with other models with the global descriptors matching, Patch-NetVLAD extracts multi-scale fusion of patch-level features that have complementary scales via an integral feature space for local descriptors matching after global descriptors matching. The local descriptors matching comes with improved accuracy. However, this operation of patch-NetVLAD is not included in the process of neural network. Therefore, it is noted that this part of calculation should not be included to compute FLOPs. But in fact, the calculation amount of Patch-NetVLAD is greater than that of VGG16-NetVALD.

GhostCNN ensures a lightweight architecture and low computational cost by replacing part of convolution operations with a series of linear transformations to generate ghost feature maps. Though the FLOPs of Ghost-dil-NetVLAD is only $1\%$ of that of VGG16-NetVLAD and Patch-NetVLAD, and the parameters is $17\%$ of theirs, Ghost-dil-NetVLAD achieves similar accuracy with them on Pitts30k test and TJU-Location test datasets. In addition, although the amount of Ghost-dil-NetVLAD’s parameters is approximately half of that
of MobileNetV3-NetVLAD and 75\% of that of Alex-NetVLAD, Ghost-dil-NetVLAD’s performance is better than theirs.


\subsubsection{Ablation Experiment for SE Module}

\begin{table}[h]
	\centering
	\caption{Recall@N for Ghost-NetVLAD with or without SE module. Note that the GhostCNN in the Ghost-NetVLAD model is not pre-trained.}
	\label{tab3}
	\setlength\tabcolsep{4.5pt}
	\begin{tabular}{ccccc}
\hline
                                &                                                               & \multicolumn{3}{c}{\textbf{Recall@N}}                             \\ \cline{3-5} 
\multirow{-2}{*}{\textbf{Dataset}}       & \multirow{-2}{*}{\textbf{Model}}                                       & 1 & 5 & 10 \\ \hline
                                & with SE                                                 & \textbf{7.7}                      & \textbf{18.16}                    & \textbf{26.39}                     \\
\multirow{-2}{*}{Pitts30k test} &  without SE & 7.14                     & 16.99                    & 24.28                     \\ \hline
	\end{tabular}
	\label{tab3}
\end{table}


The SE module adaptively re-calibrates
channel-wise feature responses by explicitly modelling inter-dependencies between channels, and then improves the useful features according to their importance and suppresses the features that are useless for the current task. This mechanism allows the model to pay more attention to the channel features with the most information to achieve high accuracy in identification and classification tasks. We performed the ablation experiment for the SE module on Pitts30k dataset to validate SE's importance. The experimental results are shown in Table 3. From the Table, the Ghost-NetVLAD model with SE module (i.e., GhostCNN with SE module) outperforms that without SE module, which proves that the SE module can promote the performance of the Ghost-NetVLAD. Similarly, in the original GhostNet \cite{han2020ghostnet}, the SE module is also used.

\subsubsection{Pre-train Experiment}

\begin{table}[t]
	\centering
	\caption{Recall@N of Ghost-NetVLAD on Pitts30k test dataset when GhostCNN is pre-trained on different datasets.}
	\label{tab3}
	\setlength\tabcolsep{4.5pt}
	\begin{tabular}{ccccc}
		\toprule
		\multirow{2}{*}{\begin{tabular}[c]{@{}c@{}}\textbf{Dataset}\end{tabular}} & \multirow{2}{*}{\textbf{Top-1 Acc.}(\%)} & \multicolumn{3}{c}{\textbf{Recall@N}} \\ \cline{3-5} 
		&                                  & \textbf{1}        & \textbf{5}       & \textbf{10}      \\ \hline
		ImageNet                                                                       & 66.2                             & 68.54    & 82.83   & 87.50   \\ \hline
		\multirow{3}{*}{Places-365}                                         & 43.84 (20 epochs)& 76.38    & 87.88   & 91.34   \\
		& 45.42 (30 epochs)& 75.95    & 87.66   & 91.34   \\
		& 46.52 (40 epochs)& 76.03    & 86.59   & 90.67   \\ 
		& 46.52 (60 epochs)& $\mathbf{76.47}$    & $\mathbf{88.57}$   & $\mathbf{91.92}$   \\ 
		\bottomrule
	\end{tabular}
	\label{tab4}
\end{table}

From Table 4, Ghost-NetVLAD achieves better performance when GhostCNN is pre-trained on the Places-365 dataset. Compared with the ImageNet dataset, the Places-365 dataset includes more building categories, like museums and palaces, etc., while a small proportion is in the ImageNet dataset. Compared with the model pre-trained on the ImageNet dataset, the model pre-trained on Places-365 improves Recall@N by at least $5\%$. To explore the impact of pre-training level on the improvement of model accuracy, we pre-train the GhostCNN on Places-365 for different epochs. Top-1 accuracy shows an increasing trend as the increasing training epochs and Recall@N of Ghost-NetVLAD gets better. For the VPR problem, a specific pre-training process can make model accuracy become better. The possible reason is that CNN architecture learns from these building images to focus on more building features to improve the model performance since most scenes are full of buildings. Therefore, we think that the targeted pre-training process can help improve model accuracy.

\subsubsection{GhostCNN with Dilated Convolution}

\begin{table}[t]
	\centering
	\caption{Recall@N of Ghost-NetVLAD with different dilated rates on Pitts30k test dataset. Dilated rate of $a$-$b$ means that the dilated rate for first four stages of GhostCNN is $a$ and $b$ in the last stage.}
	\label{tab4}
	\setlength\tabcolsep{4pt}
	\begin{tabular}{ccccc}
		\toprule
		\multirow{2}{*}{\textbf{Dataset}}       & \multirow{2}{*}{\textbf{Dilated rate}} & \multicolumn{3}{c}{\textbf{Recall@N}} \\ \cline{3-5} 
		&                                & \textbf{1}        & \textbf{5}       & \textbf{10}      \\ \midrule
		\multirow{7}{*}{Pitts30k test} & 1                      & 76.38    & 87.88   & 91.34   \\
		& 2                      & 77.32    & 88.62   & 88.62   \\
		& 3                      & 78.08    & 88.03   & 91.45   \\
		& 4                      & 77.45    & 87.72   & 91.27   \\
		& 5                      & 78.24    & 89.39   & 92.56   \\
		& 5-2                    & $\mathbf{79.45}$ & $\mathbf{89.67}$ & $\mathbf{92.80}$   \\
		& 5-3                    & 77.27    & 88.53   & 91.48   \\ \hline
		\multirow{7}{*}{TJU-Location test}       & 1                      &70.31          &76.17         &82.42         \\
		& 2                      & 69.66    & 77.47   & 82.03   \\
		& 3                      & 67.45    & 77.86   & 83.33   \\
		& 4                      & 67.06    & 77.08   & 83.72   \\
		& 5                      & 68.75    & 78.52   & 84.51   \\
		& 5-2                    & $\mathbf{70.83}$ & $\mathbf{80.08}$ & $\mathbf{85.29}$  \\
		& 5-3                    & 68.23    & 76.56   & 82.55
		\\ \hline
				\multirow{7}{*}{Tokyo 24/7}       & 1                      &34.92          &40.63         &52.06         \\
		& 2                      &35.24    &47.62   &53.33   \\
		& 3                      &$\mathbf{41.90}$    &$\mathbf{54.29}$   &$\mathbf{57.46}$   \\
		& 4                      &33.97    &47.94   &55.87   \\
		& 5                      &34.92    &46.03   &52.06   \\
		& 5-2                    &35.78    &48.57   &53.97  \\
		& 5-3                    &32.70    &50.48   &56.51\\ \bottomrule
	\end{tabular}
\end{table}

Dilated convolution can expand the receptive field to get multi-scale context information.
To further improve the accuracy of Ghost-NetVLAD, we try to apply dilated convolutions to GhostCNN on the premise of not increasing the model size and training speed. We vary the dilated rate to validate our hypothesis. From Table 5, most Recall@N results for dilated rate $>$ 1 are greater than those of the model with dilated rate $=$ 1. The results validate that the expanded receptive field could bring more features and relationships among different objects to further improve the model's performance. However, if we choose too large dilated rate, the output feature maps may be distorted.


\section{Conclusions}\label{sec:conclusions}
In this paper, to improve the original NetVLAD, we proposes a lightweight model to make a good trade-off between accuracy and model efficiency. The experimental results show that the proposed model, Ghost-dil-NetVLAD (i.e., Ghost-NetVLAD with dilated convolutions), achieves similar accuracy with VGG16-NetVLAD and outperforms other mainstream NetVLAD-based methods because dilated convolutions may expand the receptive field to get more multi-scale context information. Meanwhile, our model reduces the FLOPs and parameters of VGG16-NetVLAD by $99.04\%$ and $80.16\%$, respectively. In addition, the targeted pre-training process can help improve model accuracy efficiently. 

However, our proposed lightweight VPR model is failed to achieve the satisfactory performance in the low-light environment. Therefore, how to improve the illumination robustness of the model should be included in our future work.

\section*{Acknowledgments}
This research is supported by National Natural Science Foundation of China (61771338). 



\bibliographystyle{IEEEtran}
\bibliography{trans-refs}

\begin{thebibliography}{10}
\providecommand{\url}[1]{#1}
\csname url@samestyle\endcsname
\providecommand{\newblock}{\relax}
\providecommand{\bibinfo}[2]{#2}
\providecommand{\BIBentrySTDinterwordspacing}{\spaceskip=0pt\relax}
\providecommand{\BIBentryALTinterwordstretchfactor}{4}
\providecommand{\BIBentryALTinterwordspacing}{\spaceskip=\fontdimen2\font plus
\BIBentryALTinterwordstretchfactor\fontdimen3\font minus
  \fontdimen4\font\relax}
\providecommand{\BIBforeignlanguage}[2]{{%
\expandafter\ifx\csname l@#1\endcsname\relax
\typeout{** WARNING: IEEEtran.bst: No hyphenation pattern has been}%
\typeout{** loaded for the language `#1'. Using the pattern for}%
\typeout{** the default language instead.}%
\else
\language=\csname l@#1\endcsname
\fi
#2}}
\providecommand{\BIBdecl}{\relax}
\BIBdecl

\bibitem{lowry2015visual}
S.~Lowry, N.~S{\"u}nderhauf, P.~Newman, J.~J. Leonard, D.~Cox, P.~Corke, and
  M.~J. Milford, ``Visual place recognition: A survey,'' \emph{IEEE
  Transactions on Robotics}, vol.~32, no.~1, pp. 1--19, 2015.

\bibitem{masone2021survey}
C.~Masone and B.~Caputo, ``A survey on deep visual place recognition,''
  \emph{IEEE Access}, vol.~9, pp. 19\,516--19\,547, 2021.

\bibitem{hussain2018study}
M.~Hussain, J.~J. Bird, and D.~R. Faria, ``A study on cnn transfer learning for
  image classification,'' in \emph{UK Workshop on computational
  Intelligence}.\hskip 1em plus 0.5em minus 0.4em\relax Springer, 2018, pp.
  191--202.

\bibitem{krizhevsky2012imagenet}
A.~Krizhevsky, I.~Sutskever, and G.~E. Hinton, ``Imagenet classification with
  deep convolutional neural networks,'' \emph{Advances in neural information
  processing systems}, vol.~25, pp. 1097--1105, 2012.

\bibitem{simonyan2014very}
K.~Simonyan and A.~Zisserman, ``Very deep convolutional networks for
  large-scale image recognition,'' \emph{arXiv preprint arXiv:1409.1556}, 2014.

\bibitem{he2016deep}
K.~He, X.~Zhang, S.~Ren, and J.~Sun, ``Deep residual learning for image
  recognition,'' in \emph{Proceedings of the IEEE conference on computer vision
  and pattern recognition}, 2016, pp. 770--778.

\bibitem{han2020ghostnet}
K.~Han, Y.~Wang, Q.~Tian, J.~Guo, C.~Xu, and C.~Xu, ``Ghostnet: More features
  from cheap operations,'' in \emph{Proceedings of the IEEE/CVF Conference on
  Computer Vision and Pattern Recognition}, 2020, pp. 1580--1589.

\bibitem{ku2015discriminatively}
W.-L. Ku, H.-C. Chou, and W.-H. Peng, ``Discriminatively-learned global image
  representation using cnn as a local feature extractor for image retrieval,''
  in \emph{2015 Visual Communications and Image Processing (VCIP)}.\hskip 1em
  plus 0.5em minus 0.4em\relax IEEE, 2015, pp. 1--4.

\bibitem{arandjelovic2016netvlad}
R.~Arandjelovic, P.~Gronat, A.~Torii, T.~Pajdla, and J.~Sivic, ``Netvlad: Cnn
  architecture for weakly supervised place recognition,'' in \emph{Proceedings
  of the IEEE conference on computer vision and pattern recognition}, 2016, pp.
  5297--5307.

\bibitem{zhang2021visual}
X.~Zhang, L.~Wang, and Y.~Su, ``Visual place recognition: A survey from deep
  learning perspective,'' \emph{Pattern Recognition}, vol. 113, p. 107760,
  2021.

\bibitem{zhou2017recent}
W.~Zhou, H.~Li, and Q.~Tian, ``Recent advance in content-based image retrieval:
  A literature survey,'' \emph{arXiv preprint arXiv:1706.06064}, 2017.

\bibitem{kapoor2021state}
R.~Kapoor, D.~Sharma, and T.~Gulati, ``State of the art content based image
  retrieval techniques using deep learning: a survey,'' \emph{Multimedia Tools
  and Applications}, pp. 1--23, 2021.

\bibitem{torii2013visual}
A.~Torii, J.~Sivic, T.~Pajdla, and M.~Okutomi, ``Visual place recognition with
  repetitive structures,'' in \emph{Proceedings of the IEEE conference on
  computer vision and pattern recognition}, 2013, pp. 883--890.

\bibitem{2018GhostVLAD}
Y.~Zhong, R.~Arandjelovi, and A.~Zisserman, ``Ghostvlad for set-based face
  recognition,'' 2018.

\bibitem{9167393}
Y.~Cao, J.~Zhang, and J.~Yu, ``Image retrieval via gated multiscale netvlad for
  social media applications,'' \emph{IEEE MultiMedia}, vol.~27, no.~4, pp.
  69--78, 2020.

\bibitem{khaliq2019holistic}
A.~Khaliq, S.~Ehsan, Z.~Chen, M.~Milford, and K.~McDonald-Maier, ``A holistic
  visual place recognition approach using lightweight cnns for significant
  viewpoint and appearance changes,'' \emph{IEEE transactions on robotics},
  vol.~36, no.~2, pp. 561--569, 2019.

\bibitem{khaliq2019camal}
A.~Khaliq, S.~Ehsan, M.~Milford, and K.~McDonald-Maier, ``Camal: Context-aware
  multi-scale attention framework for lightweight visual place recognition,''
  2019.

\bibitem{babenko2014neural}
A.~Babenko, A.~Slesarev, A.~Chigorin, and V.~Lempitsky, ``Neural codes for
  image retrieval,'' in \emph{European conference on computer vision}.\hskip
  1em plus 0.5em minus 0.4em\relax Springer, 2014, pp. 584--599.

\bibitem{goodfellow2016deep}
I.~Goodfellow, Y.~Bengio, and A.~Courville, \emph{Deep learning}.\hskip 1em
  plus 0.5em minus 0.4em\relax MIT press, 2016.

\bibitem{paulin2015local}
M.~Paulin, M.~Douze, Z.~Harchaoui, J.~Mairal, F.~Perronin, and C.~Schmid,
  ``Local convolutional features with unsupervised training for image
  retrieval,'' in \emph{Proceedings of the IEEE international conference on
  computer vision}, 2015, pp. 91--99.

\bibitem{mohedano2016bags}
E.~Mohedano, K.~McGuinness, N.~E. O'Connor, A.~Salvador, F.~Marques, and
  X.~Gir{\'o}-i Nieto, ``Bags of local convolutional features for scalable
  instance search,'' in \emph{Proceedings of the 2016 ACM on International
  Conference on Multimedia Retrieval}, 2016, pp. 327--331.

\bibitem{sanchez2013image}
J.~S{\'a}nchez, F.~Perronnin, T.~Mensink, and J.~Verbeek, ``Image
  classification with the fisher vector: Theory and practice,''
  \emph{International journal of computer vision}, vol. 105, no.~3, pp.
  222--245, 2013.

\bibitem{cao2020unifying}
B.~Cao, A.~Araujo, and J.~Sim, ``Unifying deep local and global features for
  image search,'' in \emph{European Conference on Computer Vision}.\hskip 1em
  plus 0.5em minus 0.4em\relax Springer, 2020, pp. 726--743.

\bibitem{yu2019spatial}
J.~Yu, C.~Zhu, J.~Zhang, Q.~Huang, and D.~Tao, ``Spatial pyramid-enhanced
  netvlad with weighted triplet loss for place recognition,'' \emph{IEEE
  transactions on neural networks and learning systems}, vol.~31, no.~2, pp.
  661--674, 2019.

\bibitem{ge2020self}
Y.~Ge, H.~Wang, F.~Zhu, R.~Zhao, and H.~Li, ``Self-supervising fine-grained
  region similarities for large-scale image localization,'' in \emph{European
  Conference on Computer Vision}.\hskip 1em plus 0.5em minus 0.4em\relax
  Springer, 2020, pp. 369--386.

\bibitem{hausler2021patch}
S.~Hausler, S.~Garg, M.~Xu, M.~Milford, and T.~Fischer, ``Patch-netvlad:
  Multi-scale fusion of locally-global descriptors for place recognition,'' in
  \emph{Proceedings of the IEEE/CVF Conference on Computer Vision and Pattern
  Recognition}, 2021, pp. 14\,141--14\,152.

\bibitem{xu2021esa}
Y.~Xu, J.~Huang, J.~Wang, Y.~Wang, H.~Qin, and K.~Nan, ``Esa-vlad: A
  lightweight network based on second-order attention and netvlad for loop
  closure detection,'' \emph{IEEE Robotics and Automation Letters}, 2021.

\bibitem{201524}
A.~Torii, R.~Arandjelovic, J.~Sivic, M.~Okutomi, and T.~Pajdla, ``24/7 place
  recognition by view synthesis,'' in \emph{2015 IEEE Conference on Computer
  Vision and Pattern Recognition (CVPR)}, 2015.

\bibitem{howard2017mobilenets}
A.~G. Howard, M.~Zhu, B.~Chen, D.~Kalenichenko, W.~Wang, T.~Weyand,
  M.~Andreetto, and H.~Adam, ``Mobilenets: Efficient convolutional neural
  networks for mobile vision applications,'' \emph{arXiv preprint
  arXiv:1704.04861}, 2017.

\bibitem{sandler2018mobilenetv2}
M.~Sandler, A.~Howard, M.~Zhu, A.~Zhmoginov, and L.-C. Chen, ``Mobilenetv2:
  Inverted residuals and linear bottlenecks,'' in \emph{Proceedings of the IEEE
  conference on computer vision and pattern recognition}, 2018, pp. 4510--4520.

\bibitem{howard2019searching}
A.~Howard, M.~Sandler, G.~Chu, L.-C. Chen, B.~Chen, M.~Tan, W.~Wang, Y.~Zhu,
  R.~Pang, V.~Vasudevan \emph{et~al.}, ``Searching for mobilenetv3,'' in
  \emph{Proceedings of the IEEE/CVF International Conference on Computer
  Vision}, 2019, pp. 1314--1324.

\bibitem{zhang2018shufflenet}
X.~Zhang, X.~Zhou, M.~Lin, and J.~Sun, ``Shufflenet: An extremely efficient
  convolutional neural network for mobile devices,'' in \emph{Proceedings of
  the IEEE conference on computer vision and pattern recognition}, 2018, pp.
  6848--6856.

\bibitem{ma2018shufflenet}
N.~Ma, X.~Zhang, H.-T. Zheng, and J.~Sun, ``Shufflenet v2: Practical guidelines
  for efficient cnn architecture design,'' in \emph{Proceedings of the European
  conference on computer vision (ECCV)}, 2018, pp. 116--131.

\bibitem{2018Squeeze}
H.~Jie, S.~Li, and S.~Gang, ``Squeeze-and-excitation networks,'' in \emph{2018
  IEEE/CVF Conference on Computer Vision and Pattern Recognition (CVPR)}, 2018.

\bibitem{jegou2010aggregating}
H.~J{\'e}gou, M.~Douze, C.~Schmid, and P.~P{\'e}rez, ``Aggregating local
  descriptors into a compact image representation,'' in \emph{2010 IEEE
  computer society conference on computer vision and pattern
  recognition}.\hskip 1em plus 0.5em minus 0.4em\relax IEEE, 2010, pp.
  3304--3311.

\bibitem{jegou2012negative}
H.~J{\'e}gou and O.~Chum, ``Negative evidences and co-occurences in image
  retrieval: The benefit of pca and whitening,'' in \emph{European conference
  on computer vision}.\hskip 1em plus 0.5em minus 0.4em\relax Springer, 2012,
  pp. 774--787.

\bibitem{imagenet_cvpr09}
J.~Deng, W.~Dong, R.~Socher, L.-J. Li, K.~Li, and L.~Fei-Fei, ``{ImageNet: A
  Large-Scale Hierarchical Image Database},'' in \emph{CVPR09}, 2009.

\bibitem{he2019rethinking}
K.~He, R.~Girshick, and P.~Doll{\'a}r, ``Rethinking imagenet pre-training,'' in
  \emph{Proceedings of the IEEE/CVF International Conference on Computer
  Vision}, 2019, pp. 4918--4927.

\bibitem{hendrycks2019using}
D.~Hendrycks, K.~Lee, and M.~Mazeika, ``Using pre-training can improve model
  robustness and uncertainty,'' in \emph{International Conference on Machine
  Learning}.\hskip 1em plus 0.5em minus 0.4em\relax PMLR, 2019, pp. 2712--2721.

\bibitem{zhou2017places}
B.~Zhou, A.~Lapedriza, A.~Khosla, A.~Oliva, and A.~Torralba, ``Places: A 10
  million image database for scene recognition,'' \emph{IEEE transactions on
  pattern analysis and machine intelligence}, vol.~40, no.~6, pp. 1452--1464,
  2017.

\end{thebibliography}


\end{document}